# On Position Translation Vector

Yuanxin Wu[1], Zhenxiong Xiao[2]
*National University of Defense Technology, Changsha, 410073, P. R. China*

In [1], Savage described a unified mathematical framework for strapdown inertial navigation algorithm design. The commenting note [2] showed that the "velocity translation vector" (VTV) in the Savage's framework is equivalent to the dual part of an appropriate screw vector in the dual quaternion representation [3] and significantly simplified the VTV rate equation by way of the screw vector rate equation. The current note further derives a new "position translation vector" (PTV) with remarkably simpler rate equation, and proves its connections with Savage's PTV [1].

The developments in this note follow [2, 3] closely for symbol consistency. The "translation" in the context of the dual quaternion approach is a local concept between two considered frames (Footnote 4 in [3]). It could be velocity or position depending on the specific frames under investigation. Where necessary in the sequel, we discriminate the local velocity and position concepts by subscripts $v$ and $p$, respectively.

Define a new thrust position frame that is aligned with the body frame in attitude. The vector from the Earth's origin to the thrust position frame's origin is the integrated thrust velocity [3], namely, the double integrated specific force. When it comes to consider the inertial frame and the thrust position frame, we have the twist between the two frames as

$$\widehat{\boldsymbol{\omega}} = \boldsymbol{\omega} + \varepsilon\, \Delta \mathbf{t}_v^N, \tag{1}$$

where $\boldsymbol{\omega}$ denotes the gyroscope-measured body angular rate and $\Delta \mathbf{t}_v^N$ is the body-referenced thrust velocity (see Eq. (62) in [3]). Using Eq. (1) in [2], $\Delta \mathbf{t}_v^N$ can be derived as

$$\Delta \mathbf{t}_v^N = \left[ I - \frac{\sin(\sigma)}{\sigma}(\boldsymbol{\sigma}\times) + \frac{1-\cos(\sigma)}{\sigma^2}(\boldsymbol{\sigma}\times)^2 \right] \Delta \mathbf{t}_v^{N'} = \left[ I - \frac{1-\cos(\sigma)}{\sigma^2}(\boldsymbol{\sigma}\times) + \frac{1}{\sigma^2}\left(1 - \frac{\sin(\sigma)}{\sigma}\right)(\boldsymbol{\sigma}\times)^2 \right] \boldsymbol{\sigma}'_v \tag{2}$$

where $\boldsymbol{\sigma}$ and $\boldsymbol{\sigma}'_v$ are, respectively, the real part and the dual part of the dual quaternion representing the general

---

[1] Associate Professor, Department of Automatic Control, College of Mechatronics and Automation, yuanx_wu@hotmail.com.
[2] Master student, Department of Automatic Control, College of Mechatronics and Automation, eternxiao@gmail.com.

displacement from the inertial frame to the thrust frame [2, 3].

Then by mimicking the VTV derivation according to Eq. (65) in [3], the double integrated specific force (referenced in the starting frame of the update interval) is expressed as

$$\Delta \mathbf{t}_p^{N'} = \left[ I + \frac{1-\cos(\sigma)}{\sigma^2}(\boldsymbol{\sigma}\times) + \frac{1}{\sigma^2}\left(1 - \frac{\sin(\sigma)}{\sigma}\right)(\boldsymbol{\sigma}\times)^2 \right] \boldsymbol{\sigma}'_p, \tag{3}$$

where $\boldsymbol{\sigma}$ and $\boldsymbol{\sigma}'_p$ are, respectively, the real part and the dual part of the dual quaternion representing the general displacement from the inertial frame to the thrust position frame. Equation (3) is defined differently from the third equation of Eq. (8) in [1], so the dual part $\boldsymbol{\sigma}'_p$ is regarded as a new PTV. The new PTV has an obvious physical meaning in that the dual part of the associated twist in Eq. (1) is the thrust velocity $\Delta \mathbf{t}_v^N$. Comparing (3) and Eq. (8) in [1], the new PTV $\boldsymbol{\sigma}'_p$ is related to Savage's PTV $\boldsymbol{\zeta}$ by

$$F\boldsymbol{\sigma}'_p = G\boldsymbol{\zeta} \tag{4}$$

where the matrices $F$ and $G$ are as defined in Eq. (10) in [1].

Similar to Eq. (9) in [2], the rate equation for the new PTV is

$$\dot{\boldsymbol{\sigma}}'_p = \Delta \mathbf{t}_v^N + \frac{1}{2}\left(\boldsymbol{\sigma}\times\Delta\mathbf{t}_v^N + \boldsymbol{\sigma}'_p\times\boldsymbol{\omega}\right) + \frac{1}{\sigma^2}\left(1 - \frac{\sigma\sin\sigma}{2(1-\cos\sigma)}\right)\left[\boldsymbol{\sigma}\times\left(\boldsymbol{\sigma}\times\Delta\mathbf{t}_v^N\right) + \boldsymbol{\sigma}\times\left(\boldsymbol{\sigma}'_p\times\boldsymbol{\omega}\right) + \boldsymbol{\sigma}'_p\times\left(\boldsymbol{\sigma}\times\boldsymbol{\omega}\right)\right]$$
$$+ \left[\frac{\sin\sigma + \sigma}{2\sigma^3(1-\cos\sigma)} - \frac{2}{\sigma^4}\right]\left(\boldsymbol{\sigma}\cdot\boldsymbol{\sigma}'_p\right)\boldsymbol{\sigma}\times\left(\boldsymbol{\sigma}\times\boldsymbol{\omega}\right). \tag{5}$$

The above equation uses the calculated thrust velocity $\Delta \mathbf{t}_v^N$ as one of the inputs. If double integrals of the specific force or/and angular rate are available, we may express the rate equation for the new PTV as a function of VTV and employ the double integrals as inputs instead. With Eq. (2),

$$\boldsymbol{\sigma}\times\Delta\mathbf{t}_v^N = \left[\frac{\sin(\sigma)}{\sigma}(\boldsymbol{\sigma}\times) - \frac{1-\cos(\sigma)}{\sigma^2}(\boldsymbol{\sigma}\times)^2\right]\boldsymbol{\sigma}'_v,$$

$$\boldsymbol{\sigma}\times\left(\boldsymbol{\sigma}\times\Delta\mathbf{t}_v^N\right) = \left[(1-\cos(\sigma))(\boldsymbol{\sigma}\times) + \frac{\sin(\sigma)}{\sigma}(\boldsymbol{\sigma}\times)^2\right]\boldsymbol{\sigma}'_v. \tag{6}$$

Substituting Eqs. (2) and (6) into Eq. (5) yields

$$\dot{\boldsymbol{\sigma}}'_p = \boldsymbol{\sigma}'_v + \frac{1}{2}\boldsymbol{\sigma}'_p \times \boldsymbol{\omega} + \frac{1}{\sigma^2}\left(1 - \frac{\sigma \sin \sigma}{2(1-\cos\sigma)}\right)\left[\boldsymbol{\sigma} \times \left(\boldsymbol{\sigma}'_p \times \boldsymbol{\omega}\right) + \boldsymbol{\sigma}'_p \times \left(\boldsymbol{\sigma} \times \boldsymbol{\omega}\right)\right]$$
$$+ \left[\frac{\sin\sigma + \sigma}{2\sigma^3(1-\cos\sigma)} - \frac{2}{\sigma^4}\right]\left(\boldsymbol{\sigma} \cdot \boldsymbol{\sigma}'_p\right)\boldsymbol{\sigma} \times \left(\boldsymbol{\sigma} \times \boldsymbol{\omega}\right). \quad (7)$$

It appears that the new PTV is remarkably concise than Savage's PTV in terms of the rate equation (as compared to the third equation of Eq. (15) in [1]). Connections of the two rate equations are rigorously established in Appendix. For the unrealistic constant angular-rate/specific-force condition, Savage's PTV $\boldsymbol{\zeta}$ reduces to the double integral of specific force [1], while the new PTV has no such characteristic.

This note, together with [2], has successfully established the tight connections between Savage's framework [1] and the dual quaternion approach [3], which is supposed to benefit the understandings of both sides. In comparison, the powerful dual quaternion representation straightforwardly leads us to the same VTV with much simpler rate equation in [2] and a new PTV with remarkably concise rate equation in this note. It should noted that the new PTV is not an element for dual quaternion algorithm design [3]. As a matter of fact, the dual quaternion algorithm integrates the coning/sculling/scrolling algorithms all together into one unified structure, which only requires algorithm practitioners to handle the screw vector rate equation that is as simple as the differential equation of the Euler vector, in contrast to the additional manipulations of VTV and PTV in Savage's framework.

## Acknowledgments

This work was supported in part by the National Natural Science Foundation of China (61174002), the Foundation for the Author of National Excellent Doctoral Dissertation of People's Republic of China (FANEDD 200897), Program for New Century Excellent Talents in University (NCET-10-0900).

## Appendix: Connections of Two Rate Equations

The connections are built by first deriving the rate equation of Savage's PTV using Eq. (4), and then proving it is equivalent with the third equation of Eq. (15) in [1].

A. *Deriving $\dot{\boldsymbol{\zeta}}$ from Eq. (4)*

With Eq. (4), Savage's PTV is expressed as a function of the new PTV

$$\zeta = G^{-1}F\boldsymbol{\sigma}'_p \triangleq \left[I + w_1(\boldsymbol{\sigma}\times) + w_2(\boldsymbol{\sigma}\times)^2\right]\boldsymbol{\sigma}'_p \tag{8}$$

where

$$w_1 = \frac{2 - 2\cos\sigma - \sigma\sin\sigma}{2(2 + \sigma^2 - 2\cos\sigma - 2\sigma\sin\sigma)}, \quad w_2 = \frac{\left(\sigma\cos\dfrac{\sigma}{2} - 2\sin\dfrac{\sigma}{2}\right)^2}{\sigma^2(2 + \sigma^2 - 2\cos\sigma - 2\sigma\sin\sigma)} \tag{9}$$

So the rate equation is

$$\dot{\zeta} = \left[I + w_1(\boldsymbol{\sigma}\times) + w_2(\boldsymbol{\sigma}\times)^2\right]\dot{\boldsymbol{\sigma}}'_p + \left[\dot{w}_1(\boldsymbol{\sigma}\times) + w_1(\dot{\boldsymbol{\sigma}}\times) + \dot{w}_2(\boldsymbol{\sigma}\times)^2 + w_2\left((\dot{\boldsymbol{\sigma}}\times)(\boldsymbol{\sigma}\times) + (\boldsymbol{\sigma}\times)(\dot{\boldsymbol{\sigma}}\times)\right)\right]\boldsymbol{\sigma}'_p \tag{10}$$

Using the rotation vector rate equation (see e.g., Eq. (15) in [1]), $\sigma\dot{\sigma} = \boldsymbol{\sigma}\cdot\boldsymbol{\omega}$. With Eq. (7), it gives

$$\begin{aligned}
\dot{\boldsymbol{\sigma}}'_p &= \boldsymbol{\sigma}'_v + \frac{1}{2}\boldsymbol{\sigma}'_p \times \boldsymbol{\omega} + f_5\left[\boldsymbol{\sigma}\times(\boldsymbol{\sigma}'_p\times\boldsymbol{\omega}) + \boldsymbol{\sigma}'_p\times(\boldsymbol{\sigma}\times\boldsymbol{\omega})\right] + w_3(\boldsymbol{\sigma}\cdot\boldsymbol{\sigma}'_p)\boldsymbol{\sigma}\times(\boldsymbol{\sigma}\times\boldsymbol{\omega}) \\
\boldsymbol{\sigma}\times\dot{\boldsymbol{\sigma}}'_p &= \boldsymbol{\sigma}\times\boldsymbol{\sigma}'_v + \frac{1}{2}\boldsymbol{\sigma}\times(\boldsymbol{\sigma}'_p\times\boldsymbol{\omega}) + f_5\boldsymbol{\sigma}\times\left[\boldsymbol{\sigma}\times(\boldsymbol{\sigma}'_p\times\boldsymbol{\omega}) + \boldsymbol{\sigma}'_p\times(\boldsymbol{\sigma}\times\boldsymbol{\omega})\right] - \sigma^2 w_3(\boldsymbol{\sigma}\cdot\boldsymbol{\sigma}'_p)\boldsymbol{\sigma}\times\boldsymbol{\omega} \\
\boldsymbol{\sigma}\times(\boldsymbol{\sigma}\times\dot{\boldsymbol{\sigma}}'_p) &= \boldsymbol{\sigma}\times(\boldsymbol{\sigma}\times\boldsymbol{\sigma}'_v) + \frac{1}{2}\boldsymbol{\sigma}\times(\boldsymbol{\sigma}\times(\boldsymbol{\sigma}'_p\times\boldsymbol{\omega})) + f_5\left[-\sigma^2\boldsymbol{\sigma}\times(\boldsymbol{\sigma}'_p\times\boldsymbol{\omega}) - (\boldsymbol{\sigma}\cdot\boldsymbol{\sigma}'_p)\boldsymbol{\sigma}\times(\boldsymbol{\sigma}\times\boldsymbol{\omega})\right] - \sigma^2 w_3(\boldsymbol{\sigma}\cdot\boldsymbol{\sigma}'_p)\boldsymbol{\sigma}\times(\boldsymbol{\sigma}\times\boldsymbol{\omega}) \\
\dot{\boldsymbol{\sigma}}\times\boldsymbol{\sigma}'_p &= \boldsymbol{\omega}\times\boldsymbol{\sigma}'_p + \frac{1}{2}(\boldsymbol{\sigma}\times\boldsymbol{\omega})\times\boldsymbol{\sigma}'_p + f_5(\boldsymbol{\sigma}\times(\boldsymbol{\sigma}\times\boldsymbol{\omega}))\times\boldsymbol{\sigma}'_p \\
\dot{\boldsymbol{\sigma}}\times(\boldsymbol{\sigma}\times\boldsymbol{\sigma}'_p) &= \boldsymbol{\omega}\times(\boldsymbol{\sigma}\times\boldsymbol{\sigma}'_p) + \frac{1}{2}(\boldsymbol{\sigma}\times\boldsymbol{\omega})\times(\boldsymbol{\sigma}\times\boldsymbol{\sigma}'_p) + f_5(\boldsymbol{\sigma}\times(\boldsymbol{\sigma}\times\boldsymbol{\omega}))\times(\boldsymbol{\sigma}\times\boldsymbol{\sigma}'_p) \\
\boldsymbol{\sigma}\times(\dot{\boldsymbol{\sigma}}\times\boldsymbol{\sigma}'_p) &= \boldsymbol{\sigma}\times(\boldsymbol{\omega}\times\boldsymbol{\sigma}'_p) + \frac{1}{2}(\boldsymbol{\sigma}\cdot\boldsymbol{\sigma}'_p)\boldsymbol{\sigma}\times\boldsymbol{\omega} + f_5(\boldsymbol{\sigma}\cdot\boldsymbol{\sigma}'_p)\boldsymbol{\sigma}\times(\boldsymbol{\sigma}\times\boldsymbol{\omega}) \\
\dot{w}_1 &\triangleq (\boldsymbol{\sigma}\cdot\boldsymbol{\omega})w_4 \\
\dot{w}_2 &\triangleq (\boldsymbol{\sigma}\cdot\boldsymbol{\omega})w_5
\end{aligned} \tag{11}$$

where $f_i$ follow the definitions exactly in Eqs. (7) and (16) in [1] and

$$\begin{aligned}
w_3 &\triangleq \frac{\sin\sigma + \sigma}{2\sigma^3(1 - \cos\sigma)} - \frac{2}{\sigma^4} \\
w_4 &\triangleq \frac{-6\sigma + (2 + 3\sigma^2)\sin\sigma - (\sigma^2 + 2\sin\sigma - 6\sigma)\cos\sigma}{2\sigma(2 + \sigma^2 - 2\cos\sigma - 2\sigma\sin\sigma)^2} \\
w_5 &\triangleq \frac{\left(\sigma\cos\dfrac{\sigma}{2} - 2\sin\dfrac{\sigma}{2}\right)}{\sigma^4(2 + \sigma^2 - 2\cos\sigma - 2\sigma\sin\sigma)^2} \cdot \\
&\quad \left[-2\sigma(3 + \sigma^2)\cos\dfrac{\sigma}{2} + 6\sigma\cos\dfrac{3\sigma}{2} + 12\sin\dfrac{\sigma}{2} + 9\sigma^2\sin\dfrac{\sigma}{2} - \sigma^4\sin\dfrac{\sigma}{2} - 4\sin\dfrac{3\sigma}{2} + \sigma^2\sin\dfrac{3\sigma}{2}\right]
\end{aligned} \tag{12}$$

Substituting Eq. (11) and reorganizing terms, the rate equation (10) is simplified as

$$\begin{aligned}
\dot{\boldsymbol{\zeta}} =\ & \boldsymbol{\sigma}'_v \\
&+ \boldsymbol{\sigma} \times \boldsymbol{\sigma}'_v [w_1] \\
&+ \boldsymbol{\sigma} \times (\boldsymbol{\sigma} \times \boldsymbol{\sigma}'_v)[w_2] \\
&+ \boldsymbol{\sigma}'_p \times \boldsymbol{\omega} \left[ \frac{1}{2} - w_1 - \sigma^2 \left( \frac{1}{2} w_2 - f_5 w_1 \right) \right] \\
&+ \boldsymbol{\sigma} \times (\boldsymbol{\sigma}'_p \times \boldsymbol{\omega}) \left[ f_5 + \frac{1}{2} w_1 - 2 w_2 + \sigma^2 w_3 (1 - \sigma^2 w_2) \right] \\
&+ \boldsymbol{\sigma}'_p \times (\boldsymbol{\sigma} \times \boldsymbol{\omega}) \left[ f_5 - \frac{1}{2} w_1 + w_2 - \sigma^2 f_5 w_2 \right] \\
&+ (\boldsymbol{\sigma} \cdot \boldsymbol{\sigma}'_p) \boldsymbol{\sigma} \times \boldsymbol{\omega} \left[ \frac{1}{2} w_2 - 2 f_5 w_1 - \sigma^2 w_1 w_3 \right] \\
&+ (\boldsymbol{\sigma} \cdot \boldsymbol{\omega}) \boldsymbol{\sigma} \times \boldsymbol{\sigma}'_p [2 f_5 w_1 + w_4] \\
&+ (\boldsymbol{\sigma} \cdot \boldsymbol{\omega}) \boldsymbol{\sigma} \times (\boldsymbol{\sigma} \times \boldsymbol{\sigma}'_p) \left[ w_5 + f_5 w_2 + w_3 - \sigma^2 w_3 w_2 \right]
\end{aligned} \tag{13}$$

During the tedious but straightforward development of Eq. (13), the following vector equalities are frequently used

$$\begin{aligned}
\boldsymbol{\sigma} \times (\boldsymbol{\sigma} \times (\boldsymbol{\sigma}'_p \times \boldsymbol{\omega})) &= (\boldsymbol{\sigma} \cdot \boldsymbol{\omega}) \boldsymbol{\sigma} \times \boldsymbol{\sigma}'_p - (\boldsymbol{\sigma} \cdot \boldsymbol{\sigma}'_p) \boldsymbol{\sigma} \times \boldsymbol{\omega} \\
\boldsymbol{\omega} \times (\boldsymbol{\sigma} \times \boldsymbol{\sigma}'_p) &= \boldsymbol{\sigma}'_p \times (\boldsymbol{\sigma} \times \boldsymbol{\omega}) - \boldsymbol{\sigma} \times (\boldsymbol{\sigma}'_p \times \boldsymbol{\omega}) \\
\left[ \boldsymbol{\sigma}'_p \cdot (\boldsymbol{\sigma} \times \boldsymbol{\omega}) \right] \boldsymbol{\sigma} &= \boldsymbol{\sigma}'_p \times \left[ \boldsymbol{\sigma} \times (\boldsymbol{\sigma} \times \boldsymbol{\omega}) \right] + (\boldsymbol{\sigma}'_p \cdot \boldsymbol{\sigma}) \boldsymbol{\sigma} \times \boldsymbol{\omega} \\
&= (\boldsymbol{\sigma} \cdot \boldsymbol{\omega}) \boldsymbol{\sigma}'_p \times \boldsymbol{\sigma} - \sigma^2 \boldsymbol{\sigma}'_p \times \boldsymbol{\omega} + (\boldsymbol{\sigma}'_p \cdot \boldsymbol{\sigma}) \boldsymbol{\sigma} \times \boldsymbol{\omega} \\
(\boldsymbol{\sigma} \times (\boldsymbol{\sigma} \times \boldsymbol{\omega})) \times (\boldsymbol{\sigma} \times \boldsymbol{\sigma}'_p) &= (\boldsymbol{\sigma} \cdot \boldsymbol{\omega}) \boldsymbol{\sigma} \times (\boldsymbol{\sigma} \times \boldsymbol{\sigma}'_p) - \sigma^2 \boldsymbol{\omega} \times (\boldsymbol{\sigma} \times \boldsymbol{\sigma}'_p) \\
(\boldsymbol{\sigma} \cdot \boldsymbol{\sigma}'_p) \boldsymbol{\sigma} \times (\boldsymbol{\sigma} \times \boldsymbol{\omega}) &= (\boldsymbol{\sigma} \cdot \boldsymbol{\omega}) \boldsymbol{\sigma} \times (\boldsymbol{\sigma} \times \boldsymbol{\sigma}'_p) + \sigma^2 \boldsymbol{\sigma} \times (\boldsymbol{\sigma}'_p \times \boldsymbol{\omega})
\end{aligned} \tag{14}$$

*B. Reduction of $\dot{\boldsymbol{\zeta}}$, Third Equation of Eq. (15) in [1]*

With the differential equation of the rotation vector in Eq. (15) in [1],

$$\dot{\boldsymbol{\phi}} = \boldsymbol{\omega} + \frac{1}{2}\boldsymbol{\phi}\times\boldsymbol{\omega} + f_5\boldsymbol{\phi}\times(\boldsymbol{\phi}\times\boldsymbol{\omega})$$

$$\dot{\boldsymbol{\phi}}\times\boldsymbol{\zeta} = (1-f_5\phi^2)\boldsymbol{\omega}\times\boldsymbol{\zeta} + \frac{1}{2}(\boldsymbol{\phi}\times\boldsymbol{\omega})\times\boldsymbol{\zeta} + f_5(\boldsymbol{\phi}\cdot\boldsymbol{\omega})\boldsymbol{\phi}\times\boldsymbol{\zeta}$$

$$\boldsymbol{\phi}\times(\dot{\boldsymbol{\phi}}\times\boldsymbol{\zeta}) = (1-f_5\phi^2)\boldsymbol{\phi}\times(\boldsymbol{\omega}\times\boldsymbol{\zeta}) + \frac{1}{2}(\boldsymbol{\phi}\cdot\boldsymbol{\zeta})\boldsymbol{\phi}\times\boldsymbol{\omega} + f_5(\boldsymbol{\phi}\cdot\boldsymbol{\omega})\boldsymbol{\phi}\times(\boldsymbol{\phi}\times\boldsymbol{\zeta})$$

$$\boldsymbol{\phi}\times[\boldsymbol{\phi}\times(\dot{\boldsymbol{\phi}}\times\boldsymbol{\zeta})] = (1-f_5\phi^2)[(\boldsymbol{\phi}\cdot\boldsymbol{\zeta})\boldsymbol{\phi}\times\boldsymbol{\omega} - (\boldsymbol{\phi}\cdot\boldsymbol{\omega})\boldsymbol{\phi}\times\boldsymbol{\zeta}] + \frac{1}{2}(\boldsymbol{\phi}\cdot\boldsymbol{\zeta})\boldsymbol{\phi}\times(\boldsymbol{\phi}\times\boldsymbol{\omega}) - \phi^2 f_5(\boldsymbol{\phi}\cdot\boldsymbol{\omega})\boldsymbol{\phi}\times\boldsymbol{\zeta}$$

$$\boldsymbol{\phi}\times\dot{\boldsymbol{\phi}} = \boldsymbol{\phi}\times\boldsymbol{\omega} + \frac{1}{2}\boldsymbol{\phi}\times(\boldsymbol{\phi}\times\boldsymbol{\omega}) - \phi^2 f_5 \boldsymbol{\phi}\times\boldsymbol{\omega}$$

$$(\boldsymbol{\phi}\times\dot{\boldsymbol{\phi}})\times\boldsymbol{\zeta} = (1-f_5\phi^2)(\boldsymbol{\phi}\times\boldsymbol{\omega})\times\boldsymbol{\zeta} + \frac{1}{2}[(\boldsymbol{\phi}\cdot\boldsymbol{\omega})\boldsymbol{\phi}\times\boldsymbol{\zeta} - \phi^2\boldsymbol{\omega}\times\boldsymbol{\zeta}]$$

$$\boldsymbol{\phi}\times[(\boldsymbol{\phi}\times\dot{\boldsymbol{\phi}})\times\boldsymbol{\zeta}] = (1-f_5\phi^2)(\boldsymbol{\phi}\cdot\boldsymbol{\zeta})\boldsymbol{\phi}\times\boldsymbol{\omega} + \frac{1}{2}[(\boldsymbol{\phi}\cdot\boldsymbol{\omega})\boldsymbol{\phi}\times(\boldsymbol{\phi}\times\boldsymbol{\zeta}) - \phi^2\boldsymbol{\phi}\times(\boldsymbol{\omega}\times\boldsymbol{\zeta})]$$

$$\boldsymbol{\phi}\times\{\boldsymbol{\phi}\times[(\boldsymbol{\phi}\times\dot{\boldsymbol{\phi}})\times\boldsymbol{\zeta}]\} = (1-f_5\phi^2)(\boldsymbol{\phi}\cdot\boldsymbol{\zeta})\boldsymbol{\phi}\times(\boldsymbol{\phi}\times\boldsymbol{\omega}) + \frac{1}{2}\{-(\boldsymbol{\phi}\cdot\boldsymbol{\omega})\phi^2\boldsymbol{\phi}\times\boldsymbol{\zeta} - \phi^2[(\boldsymbol{\phi}\cdot\boldsymbol{\zeta})\boldsymbol{\phi}\times\boldsymbol{\omega} - (\boldsymbol{\phi}\cdot\boldsymbol{\omega})\boldsymbol{\phi}\times\boldsymbol{\zeta}]\}$$

(15)

Substituting into Eq. (15) in [1] and reorganizing terms, the rate equation for Savage's PTV is reduced to

$$\dot{\boldsymbol{\zeta}} \triangleq \boldsymbol{\eta} + g_1\boldsymbol{\phi}\times\boldsymbol{\eta} + g_2\boldsymbol{\phi}\times(\boldsymbol{\phi}\times\boldsymbol{\eta}) + g_3\boldsymbol{\omega}\times\boldsymbol{\zeta} + g_4(\boldsymbol{\phi}\times\boldsymbol{\omega})\times\boldsymbol{\zeta} + g_5(\boldsymbol{\phi}\cdot\boldsymbol{\omega})\boldsymbol{\phi}\times\boldsymbol{\zeta}$$
$$+ g_6(\boldsymbol{\phi}\cdot\boldsymbol{\zeta})\boldsymbol{\phi}\times\boldsymbol{\omega} + g_7\boldsymbol{\phi}\times(\boldsymbol{\omega}\times\boldsymbol{\zeta}) + g_8(\boldsymbol{\phi}\cdot\boldsymbol{\omega})\boldsymbol{\phi}\times(\boldsymbol{\phi}\times\boldsymbol{\zeta}) + g_9(\boldsymbol{\phi}\cdot\boldsymbol{\zeta})\boldsymbol{\phi}\times(\boldsymbol{\phi}\times\boldsymbol{\omega})$$

(16)

where $g_i$ are defined as

$$g_1 = \frac{1}{6} - f_{11}\phi^2,\ g_2 = f_9,\ g_3 = -\left(\frac{1}{3} - f_{12}\phi^2\right)(1-f_5\phi^2) - f_4\phi^2,$$

$$g_4 = 2f_4(1-f_5\phi^2) - \frac{1}{2}\left(\frac{1}{3} - f_{12}\phi^2\right),\ g_5 = f_4 - f_5\left(\frac{1}{3} - f_{12}\phi^2\right) - f_{13}(1-f_5\phi^2) - f_{13}f_5\phi^2 + f_{14},$$

$$g_6 = \frac{1}{2}(f_{16}\phi^2 - 2f_9) - f_{10}(1-f_5\phi^2) + f_{13}(1-f_5\phi^2) - \frac{1}{2}f_{15}\phi^2,\ g_7 = (f_{16}\phi^2 - 2f_9)(1-f_5\phi^2) + \frac{1}{2}f_{10}\phi^2,$$

$$g_8 = (f_{16}\phi^2 - 2f_9)f_5 - \frac{1}{2}f_{10} - f_{17},\ g_9 = \frac{1}{2}f_{13} + f_{15}(1-f_5\phi^2)$$

(17)

To be consistent in symbol notations, replacing $\boldsymbol{\phi}$, $\boldsymbol{\eta}$ and $\boldsymbol{\zeta}$ by $\boldsymbol{\sigma}$, $\boldsymbol{\sigma}'_v$ and $\left[I + w_1(\boldsymbol{\sigma}\times) + w_2(\boldsymbol{\sigma}\times)^2\right]\boldsymbol{\sigma}'_p$, respectively, the vector terms in Eq. (16) are transformed to

$$\zeta = \left[ I + w_1 \boldsymbol{\sigma} \times + w_2 (\boldsymbol{\sigma} \times)^2 \right] \boldsymbol{\sigma}'_p$$

$$\boldsymbol{\phi} \cdot \zeta = \boldsymbol{\sigma} \cdot \boldsymbol{\sigma}'_p$$

$$\boldsymbol{\eta} = \boldsymbol{\sigma}'_v$$

$$\boldsymbol{\phi} \times \boldsymbol{\eta} = \boldsymbol{\sigma} \times \boldsymbol{\sigma}'_v$$

$$\boldsymbol{\phi} \times (\boldsymbol{\phi} \times \boldsymbol{\eta}) = \boldsymbol{\sigma} \times (\boldsymbol{\sigma} \times \boldsymbol{\sigma}'_v)$$

$$\boldsymbol{\omega} \times \zeta = (1 - \sigma^2 w_2) \boldsymbol{\omega} \times \boldsymbol{\sigma}'_p + w_1 \boldsymbol{\omega} \times (\boldsymbol{\sigma} \times \boldsymbol{\sigma}'_p) + w_2 (\boldsymbol{\sigma} \cdot \boldsymbol{\sigma}'_p) \boldsymbol{\omega} \times \boldsymbol{\sigma}$$

$$\boldsymbol{\phi} \times (\boldsymbol{\omega} \times \zeta) = (1 - \sigma^2 w_2) \boldsymbol{\sigma} \times (\boldsymbol{\omega} \times \boldsymbol{\sigma}'_p) - w_1 (\boldsymbol{\sigma} \cdot \boldsymbol{\omega}) \boldsymbol{\sigma} \times \boldsymbol{\sigma}'_p + w_2 (\boldsymbol{\sigma} \cdot \boldsymbol{\sigma}'_p) \boldsymbol{\sigma} \times (\boldsymbol{\omega} \times \boldsymbol{\sigma})$$

$$(\boldsymbol{\phi} \cdot \zeta) \boldsymbol{\phi} \times \boldsymbol{\omega} = (\boldsymbol{\sigma} \cdot \boldsymbol{\sigma}'_p) \boldsymbol{\sigma} \times \boldsymbol{\omega}$$

$$(\boldsymbol{\phi} \cdot \zeta) \boldsymbol{\phi} \times (\boldsymbol{\phi} \times \boldsymbol{\omega}) = (\boldsymbol{\sigma} \cdot \boldsymbol{\sigma}'_p) \boldsymbol{\sigma} \times (\boldsymbol{\sigma} \times \boldsymbol{\omega})$$

$$(\boldsymbol{\phi} \times \boldsymbol{\omega}) \times \zeta = (1 - \sigma^2 w_2)(\boldsymbol{\sigma} \times \boldsymbol{\omega}) \times \boldsymbol{\sigma}'_p + w_1 \left[ \boldsymbol{\sigma}'_p \cdot (\boldsymbol{\sigma} \times \boldsymbol{\omega}) \right] \boldsymbol{\sigma} + w_2 (\boldsymbol{\sigma} \cdot \boldsymbol{\sigma}'_p)(\boldsymbol{\sigma} \times \boldsymbol{\omega}) \times \boldsymbol{\sigma}$$

$$(\boldsymbol{\phi} \cdot \boldsymbol{\omega}) \boldsymbol{\phi} \times \zeta = (1 - \sigma^2 w_2)(\boldsymbol{\sigma} \cdot \boldsymbol{\omega}) \boldsymbol{\sigma} \times \boldsymbol{\sigma}'_p + w_1 (\boldsymbol{\sigma} \cdot \boldsymbol{\omega}) \boldsymbol{\sigma} \times (\boldsymbol{\sigma} \times \boldsymbol{\sigma}'_p)$$

$$(\boldsymbol{\phi} \cdot \boldsymbol{\omega}) \boldsymbol{\phi} \times (\boldsymbol{\phi} \times \zeta) = (1 - \sigma^2 w_2)(\boldsymbol{\sigma} \cdot \boldsymbol{\omega}) \boldsymbol{\sigma} \times (\boldsymbol{\sigma} \times \boldsymbol{\sigma}'_p) - \sigma^2 w_1 (\boldsymbol{\sigma} \cdot \boldsymbol{\omega}) \boldsymbol{\sigma} \times \boldsymbol{\sigma}'_p \tag{18}$$

Substituting into (16) and using (14), we have

$$\begin{aligned}
\dot{\zeta} &= \boldsymbol{\sigma}'_v \\
&+ \boldsymbol{\sigma} \times \boldsymbol{\sigma}'_v \left[ g_1 \right] \\
&+ \boldsymbol{\sigma} \times (\boldsymbol{\sigma} \times \boldsymbol{\sigma}'_v) \left[ g_2 \right] \\
&+ \boldsymbol{\sigma}'_p \times \boldsymbol{\omega} \left[ -(1 - \sigma^2 w_2) g_3 - \sigma^2 w_1 g_4 \right] \\
&+ \boldsymbol{\sigma} \times (\boldsymbol{\sigma}'_p \times \boldsymbol{\omega}) \left[ \sigma^2 (-w_2 g_4 - w_2 g_7 + g_9) - w_1 g_3 - (1 - \sigma^2 w_2) g_7 \right] \\
&+ \boldsymbol{\sigma}'_p \times (\boldsymbol{\sigma} \times \boldsymbol{\omega}) \left[ w_1 g_3 - (1 - \sigma^2 w_2) g_4 \right] \\
&+ (\boldsymbol{\sigma} \cdot \boldsymbol{\sigma}'_p) \boldsymbol{\sigma} \times \boldsymbol{\omega} \left[ g_6 - w_2 g_3 + w_1 g_4 \right] \\
&+ (\boldsymbol{\sigma} \cdot \boldsymbol{\omega}) \boldsymbol{\sigma} \times \boldsymbol{\sigma}'_p \left[ (1 - \sigma^2 w_2) g_5 - w_1 g_7 - \sigma^2 w_1 g_8 - w_1 g_4 \right] \\
&+ (\boldsymbol{\sigma} \cdot \boldsymbol{\omega}) \boldsymbol{\sigma} \times (\boldsymbol{\sigma} \times \boldsymbol{\sigma}'_p) \left[ w_1 g_5 + (1 - \sigma^2 w_2) g_8 - w_2 g_4 - w_2 g_7 + g_9 \right]
\end{aligned} \tag{19}$$

It can be verified, for example by the help of Mathematica, that the coefficients of vector terms above are exactly those in Eq. (13) correspondingly. In other words, the new PTV is truly related to Savage's PTV by Eq. (4).